\documentclass[11pt]{article}
\usepackage{geometry}
\usepackage{coling2020}
\usepackage{times}
\usepackage{url}
\usepackage{graphicx}
\usepackage{latexsym}
\usepackage{microtype}
\usepackage{booktabs}
\usepackage{multirow} 

\colingfinalcopy % Uncomment this line for all SemEval submissions

\title{IUST at SemEval-2020 Task 9: Sentiment Analysis for Code-Mixed Social Media Text using Deep Neural Networks and Linear Baselines}

\author{Soroush Javdan, Taha Shangipour ataei and Behrouz Minaei-Bidgoli\\
	Computer Engineering Department \\
	Iran University of Science and Technology \\
	Tehran, Iran \\
	{\tt soroush\_javdan,taha\_shangipour@comp.iust.ac.ir} \\
	{\tt b\_minaei@iust.ac.ir} \\}

\date{}

\begin{document}
\maketitle
\begin{abstract}
	Sentiment Analysis is a well-studied field of Natural Language Processing. However, the rapid growth of social media and noisy content within them poses significant challenges in addressing this problem with well-established methods and tools. One of these challenges is code-mixing, which means using different languages to convey thoughts in social media texts. Our group, with the name of IUST(username: TAHA), participated at the SemEval-2020 shared task 9 on Sentiment Analysis for Code-Mixed Social Media Text, and we have attempted to develop a system to predict the sentiment of a given code-mixed tweet. We used different preprocessing techniques and proposed to use different methods that vary from NBSVM to more complicated deep neural network models. Our best performing method obtains an F1 score of 0.751 for the Spanish-English sub-task and 0.706 over the Hindi-English sub-task.
\end{abstract}

\section{Introduction}
\label{intro}

\blfootnote{
	%
	% for review submission
	%
	%
	% % final paper: en-uk version 
	%
	% \hspace{-0.65cm}  % space normally used by the marker
	% This work is licensed under a Creative Commons 
	% Attribution 4.0 International Licence.
	% Licence details:
	% \url{http://creativecommons.org/licenses/by/4.0/}.
	% 
	% % final paper: en-us version 
	
	\hspace{-0.65cm}  % space normally used by the marker
	This work is licensed under a Creative Commons 
	Attribution 4.0 International License.
	License details:
	\url{http://creativecommons.org/licenses/by/4.0/}.
}

Sentiment Analysis identification is a sub-field of natural language processing that explores the automatic inference for fine-grained opinion polarity of textual data. The recent growth of social media, review forums and text messaging platforms create a surge in the amount of user-generated textual data, and so increased the urgent need for automatic opinion extraction. However this evolution created many opportunities for language technology and researchers, it provided verity of new challenges, namely, spelling errors, creative invented spelling ("iz" for "is"), abbreviation ("LOL" for "laugh out loud"), Meta tags(Hashtags) and code-mixing \cite{das2014identifying}. Non-English speakers frequently use multiple languages to express their feelings, which they know as code-mixing.  Speakers or writers tend to shift from one language to another either to express their feelings adequately, to show like-mindedness with a group, to distinguish oneself, to discuss a specific topic, or to look impressive to their audience. The SemEval-2020 shared task 9 \cite{patwa2020sentimix}, which is part of the SemEval-2020 workshop, focused on coping with this challenge. This task focus is automatic sentiment analysis in a code-mixed social media text. The task consists of two subtasks, Spanish-English(Spanglish) and Hindi-English(Hinglish) code-mixing, with over 30 groups who participate in each sub-task. 

This article presents a system that we have implemented for predicting the sentiment of a given code-mixed tweet. The system was developed for both subtasks. Table 1 shows the results that we achieved with our system in the SemEval-2020 competitions. To create a highly accurate classifier, we tested different methods that varied from linear (NBSVM with n-grams) and different deep learning architectures (LSTM, BiLSTM, and combinations). Also, we tested a new architecture based on Wang's \shortcite{wang-etal-2019-ynuwb}  work on offensive language detection, who used four Convolutional Neural Networks(CNN) with different window sizes(1,2,3,4) and k max-pooling ahead of them. However, the combination of Term Frequency-Inverse Document Frequency(TF-IDF) embedding and NBSVM yield the best result.

\begin{table}[t!]
	\begin{center}
		\begin{tabular}{|c|cl|}
			\hline \bf Competition & \bf Placement & \\ \hline
			English-Hinish & 6\textsuperscript{th} &  \\
			English-Spanish & 7\textsuperscript{th} &  \\
			\hline
		\end{tabular}
	\end{center}
	\caption{\label{placement-table} SemEval-2020 results. }
\end{table}

\section{Related Work}

Sentiment analysis is a well-established and well-studied topic in social media analysis and natural language processing(NLP); however, most of the effort has been applied to addressing this problem in the monolingual context. Since English is an international language and is the second most spoken language globally, it is most likely to be code-mixed with other languages. Hence, most of the works in the code-mixed criteria are focused on the combination of English with other languages.

Bhargava \shortcite{bhargava2016sentiment} had divided the Code-mixed Sentiment Analysis of English and Indian languages problem, into the Language Identification and Sentiment Analysis sub-tasks. First, they used word-level language labeling and then applied SentiWordNet to analyze the semantic of each word. Overal sentiments of the sentence had obtained from all the words' sentiments. Ghosh \shortcite{ghosh2017sentiment} adopted a multi-layer perceptron with varieties of extracted features such as Part-Of-Speech Tags, frequency of code switches, the density of curse words, and the number of slang words. Parvalika \shortcite{pravalika2017domain} proposed a method based on a lexicon-based and machine learning approaches to address Hindi-English Sentiment Analysis. In the lexicon-based method, they used dictionaries of words annotated with their polarity to classify text. In the machine learning-based part, first, they did data cleaning and replace slang words with their associated meaning. Then, Naive Bayes(NB), Support Vector Machine(SVM), Decision Tree(DT), Random Tree(RT), Multilayer Perceptron were applied for text classification purposes. Prabhu \shortcite{prabhu2016towards} addressed the same problem with sub-word level representations in LSTM architecture. They applied character-level embedding followed by convolution with a filter length of 3 in order to obtain sub-word level representations. LSTM was utilized to model the relationships between obtained features. The LSTM output feeds to a fully connected layer to calculate the final label. Lal \shortcite{lal2019mixing} proposed a model that consists of three different components. The first component prepared the sub-word level representations. Then the output was feed to the second component, which comprises two different BiLSTM. One BiLSTM aims to learn a representation for the overall sentiment of the sentence, and the other one, which is followed by an affixed attention mechanism, enables the selection of subwords that contribute the most towards the input text's sentiment. Finally, the concatenation of two BiLSTM aligns with linguistic features feed to a fully connected layer to predict sentence polarity. In order to address the Bambara-French code-mixed sentiment analysis problem in Facebook comments, Konate \shortcite{konate2018sentiment} employed a variety of deep learning methods, including LSTM, BiLSTM, CNN, and their combinations. Character and word embedding produced from fixed indexes were used due to the lake of pre-trained word vectors for the Bambara language. Vilares \shortcite{vilares2015sentiment} investigated sentiment analysis for English and Spanish language separately and their combination on twitter. They extracted a set of linguistic features. They also make use of dependency parsers to achieve syntactic features. They developed an utterly monolingual model, a pipeline that adopted language identification techniques to discover the language of texts, and a multilingual model trained on Spanish-English texts. For the word embedding phase in resolving the Spanish-English code-mixed sentiment analysis problem, Yadav \shortcite{yadav2020unsupervised} used various methods. In the first approach, they separately learn word embedding for each language and then used supervised and unsupervised approaches to learn a transformation matrix to map representation from one language to the representation of the other language. In the second method, they trained word, and subword representation on corpus contains text from both languages and their combination. Last, they adopted a Machine Translation approach to learning the cross-lingual sentence-level representation. They achieved the best result with the combination of subword representation and LSTM.

\section{Data}

\begin{table*}[t]
	\begin{center}
		\begin{tabular}{|c|c|cl|}
			\hline \bf Labels & \bf Hindi-English (\%) & \bf Spanish-English (\%) & \\ \hline
			English words only & 41.48 & 32.41 &  \\
			Hindi/Spanish words only & 58.52 & 67.59 & \\
			\hline
		\end{tabular}
	\end{center}
	\caption{\label{language-labels-table} A brief description of language labels distribution for the Hindi-English and Spanish-English datasets. }
\end{table*}

Two corpora were used in Code-Mixed Sentiment Analysis
shared task. The Spanish-English dataset consists of 15000 data samples for the train and 3789 for the test. On the other hand, the Hindi-English dataset contains total 15131 data samples for the train and 3000 for the test set. The sentiment labels are positive, negative, or neutral. Besides the sentiment labels, the language labels at the word level have been provided. The distribution of both language labels for each dataset is shown in Table~\ref{language-labels-table}. The distribution of Hindi-English indicates that using English words is a more common practice in India.

\begin{figure*}[tph!]
	\centerline{\includegraphics[totalheight=5cm]{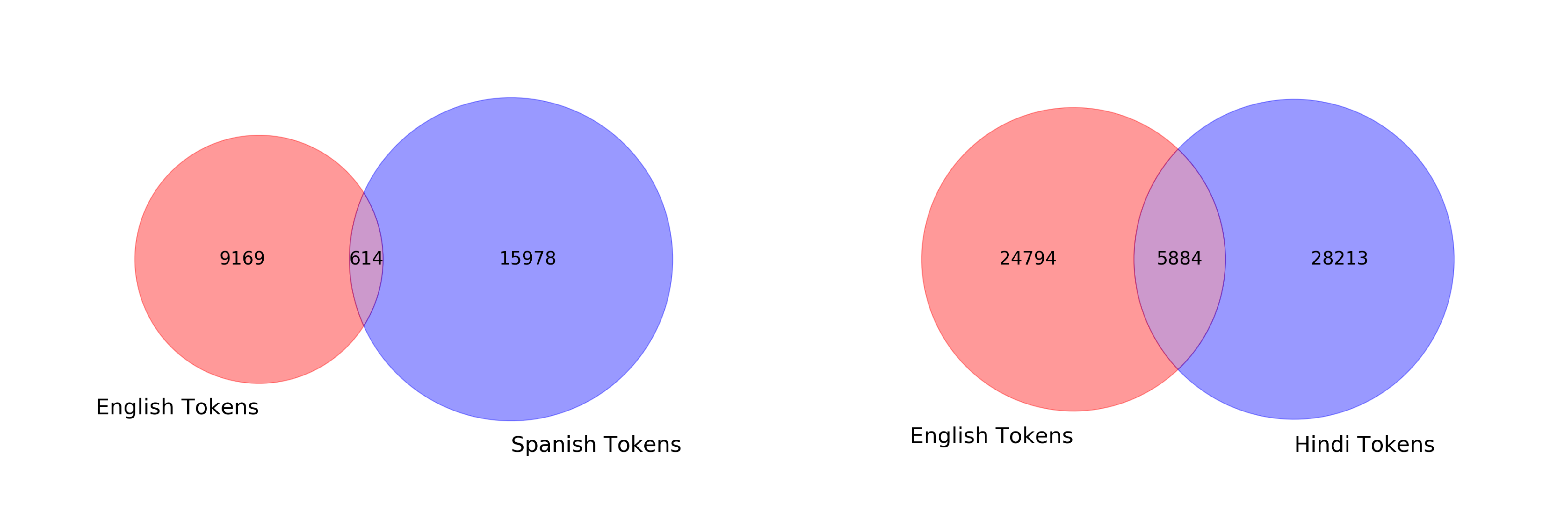}}
	\caption{Vocabulary overlap for language pairs for Spanish-English and Hindi-English datasets.}
	\label{fig:overlap}
\end{figure*}

Figure~\ref{fig:overlap} shows the overlap between the unique tokens between the first and second languages. There are 614 words in the Spanish-English dataset, which have the same spelling in both languages. This number is much higher for the Hindi-English dataset. It contains 5884 words with similar spelling for both Hindi and English languages, and approximately, there is an 11\% overlap between English and Hindi vocabulary.

\section{Methodology}

In this section, we describe the techniques and models that we applied to address the code-mixed Sentiment Analysis problem.

\subsection{Preprocessing}
 
{\bf Tokenization:}
Since social media content is noisy and contains user mentions and hashtags, the tokenization process can be turned to a challenging task. We utilized Keras\footnote{https://keras.io/preprocessing/text/ - tokenizer} and nltk's\footnote{http://nltk.org/} tweet tokenizers.

{\bf Hashtag segmentation:}
Hashtag is metadata, helping people find messages with the same topic within social media. Hashtag start with a number sign, \#, and usually follows by a combination of words. We removed \# sign and segmented the words in hashtags.

{\bf Removing URLs:}
we removed all of the URLs in data.

{\bf Misc.:}
We lowercased all the letters of the training and test data.

\subsection{Models}

{\bf NBSVM:}
We employed the NBSVM \cite{wang-manning-2012-baselines} model, a combination of Naïve Bayes and linear models such as Support Vector Machine(or logistic regression). The calculation ratio of log Naïve Bayes used as a feature for Support Vector Machine. Notwithstanding its simplicity, NBSVM methods have been shown to be both fast and robust across a wide variety of different text classification tasks. We fed the TF-IDF matrix with character n-gram features to model as input. We applied this method over both of the Hindi-English and Spanish-English datasets. 

{\bf SVM:}
We used Support Vector Machine with the TF-IDF matrix with character n-gram made from data as input features.

{\bf XGboost:}
We utilized XGBoost \cite{chen2016xgboost}, an optimized and distributed version of the gradient boosting model, and the TF-IDF matrix with character n-gram made from data.

{\bf Bi-GRU-CNN+BiLSTM-CNNL:}
Convolutional neural network is employed in text classification tasks as it is suitable for detecting patterns, and also it can detect different patterns by changing kernel sizes in different positions. The recurrent neural network is a sequence of linked network blocks that each block passes a message to its successor, this feature enables the network to demonstrate dynamic temporal behavior for a time sequence and capture sequential data. ata. For this model, we also used fastText \cite{mikolov2018advances} word embedding to extract word vector representation of data. The excellent characteristic of fastText is that it is derived from character n-gram that makes it convenient for social media's noisy content. We employ a neural network architecture built on top of a fastText English embedding with 300 dimensions. Then, the network splits into two parallel sections; the first section combines a bidirectional gated recurrent unit (GRU) with 128 hidden units and a convolutional layer with a kernel size of 2 and 64 hidden units. The second section combines a bidirectional long short term memory (LSTM) with 128 hidden units and a convolutional layer with a kernel size of 2 and 64 hidden units.  Finally, we concatenate global max-pooling and global average pooling layers over parallel parts and send them to a dense layer followed with the softmax layer for classification purposes.

{\bf DistilBERT-LR:}
Bidirectional Encoder Representation from Transformer (BERT) \cite{devlin2018bert} released by Google research team, achieved state of the art in many NLP tasks. Moreover, since it has more than 100 million parameters, the fine-tuning procedure requires GPUs and may take a long time. Thus, we use a distilled version of BERT introduced in Sanh \shortcite{sanh2019distilbert}. They reduced its parameters by 40\% and retained its language understanding by 97\%. We used the last layer output of DistilBERT and fed it to the logistic regression classifier. We tested both base-cased and multilingual pre-trained BERT. However, the result of multilingual BERT was slightly better.

{\bf CNN-GloVe+fastText:}
Same as Wang \shortcite{wang-etal-2019-ynuwb}, we used the average of fastText and GloVe \cite{pennington2014glove} embeddings with 300 dimensions as an input of a neural network which consists of four parallel convolutional layers with 64 hidden units and kernel sizes of 1,2,3 and 4 that their output was fed to a specific k max-pooling layers and the final output was used for prediction.

{\bf CNN-fastText:}
Like the previous model, we employed fastText embedding with 300 dimensions as an input of a deep neural network, which consists of four parallel convolutional layers with 128 hidden units and kernel sizes of 3, 4, 5 and 6. The batch normalization was applied after each convolutional layer, and the dropout and global max-pooling layers were used after that. Finally, the concatenation of four outputs is used for prediction.

\section{Results}

\begin{table}[h!]
	\begin{center}
		\begin{tabular}{lcccc}
			&   & \multicolumn{2}{c}{f1 score}  \\
			\textbf{Method}   & \textbf{Negative} & \textbf{Neutral} & \textbf{Positive} & \textbf{Macro Average} \\
			\hline
			\textbf{NBSVM ngram range 2 and 6} & \textbf{0.72} & \textbf{0.65} & \textbf{0.76} & \textbf{0.71} \\
			\hline
			XGB ngram range 2 and 5  & 0.59 & 0.59 & 0.67 & 0.62  \\
			\hline
			CNN-Fasttext  & 0.69 & 0.59 & 0.55 & 0.66  \\
			\hline
			DistilBERT + LR & 0.64 & 0.50 & 0.70 & 0.61 \\
			\hline
			CNN-GloVe+Fasttext & 0.70 & 0.56 & 0.72 & 0.66 \\
		\end{tabular}
		\caption{\label{result_hindi}Models performance over Hindi-English dataset}
	\end{center}
\end{table}

For all models, we used inputs with and without preprocessing. Also, we experiment with different parameter and reports the set of parameters with the best performance. Moreover, we tested all the models, as mentioned earlier, for both Hindi-English and Spanish-English datasets. The performance of NBSVM was quite impressive. We used the character n-gram with a range of 2 to 6 and the TFIDT matrix. The language independence characteristic of this model makes it a strong candidate for the code-mixed classification tasks. Our experiment with the DistilBERT model was not satisfying. However, multilingual BERT performed slightly better. The combination of CNN, fastText, and GloVe was not as expected. The poor performance was as a result of using only English embedding and ignoring the other language. The complete results of our tested models for Hindi-English and Spanish-English are shown in Table~\ref{result_hindi} and ~\ref{result_spanish}, respectively.

\begin{table}[h!]
	\begin{center}
		\begin{tabular}{lcccc}
			&   & \multicolumn{2}{c}{f1 score}  \\
			\textbf{Method}   & \textbf{Negative} & \textbf{Neutral} & \textbf{Positive} & \textbf{Macro Average} \\
			\hline
			\textbf{NBSVM ngram range 2 and 6} & 0.336 & \textbf{0.164} & \textbf{0.841} & \textbf{0.750} \\
			\hline
			Bi-GRU-CNN+BiLSTM-CNN   & \textbf{0.372} & 0.158 & 0.837 & 0.736  \\
			\hline
			DistilBERT-LR (BERT-base-cased)  &  &  &  & 0.730  \\
			\hline
			DistilBERT-LR (BERT-multilingual) &  &  &  & 0.734 \\
			\hline
			CNN-Fasttext &  &  &  & 0.682 \\
			\hline
			SVM & & & & 0.671
		\end{tabular}
		\caption{\label{result_spanish}Models performance over Spanish-English dataset}
	\end{center}
\end{table}

\section{Conclusion}

Our proposed methods ranked 6 out of 62 groups for the Hindi-English dataset and ranked 7 out of 29 for the Spanish-English dataset. This result shows the strength of the combination of NBSVM and  TF-IDF as a language independence Model. Also, our experiment shows the substantial adverse effect of ignoring one of the language's representation. For future work, we are going to utilize both languages pre-trained embedding. Averaging is one way to combine both embeddings. However, the different output space of both embedding could be challenging. Also, the attention mechanism could be used to give the model the ability to decide the importance of each language representation within each sentence.
\bibliographystyle{coling}
\bibliography{semeval2020}

\end{document}